\documentclass[letterpaper, 10 pt, conference]{ieeeconf}  %

\IEEEoverridecommandlockouts                              %

\usepackage{graphics} %
\usepackage{epsfig} %
\usepackage{times} %
\usepackage{amsmath} %
\usepackage{amssymb}  %
\usepackage{cite}
\usepackage[font=footnotesize]{caption}
\usepackage{pifont}%
\newcommand{\cmark}{\ding{51}}%
\newcommand{\xmark}{\ding{55}}%
\usepackage{multirow}
\usepackage{hyperref}
\hypersetup{
    colorlinks=true,
    linkcolor=blue,
    filecolor=magenta,      
    urlcolor=cyan,
    pdftitle={Overleaf Example},
    pdfpagemode=FullScreen,
    }
\usepackage{bm}

\usepackage[dvipsnames]{xcolor}

\title{\Large \bf
TURTLMap: Real-time Localization and Dense Mapping of Low-texture Underwater Environments with a Low-cost Unmanned Underwater Vehicle
}

 \author{%
Jingyu Song$^{*1}$, Onur Bagoren$^{*1}$, Razan Andigani$^{1}$, Advaith Sethuraman$^{1}$, and Katherine A. Skinner$^{1}$
    \thanks{This work relates to Department of Navy award N00014-21-1-2149 issued by the Office of Naval Research.}
    \thanks{$^*$denotes equal contribution}
    \thanks{$^1$J. Song, O. Bagoren, R. Andigani, A. Sethuraman, and K.A. Skinner are with the Department of Robotics, University of Michigan, Ann Arbor, MI 48109 USA. Corresponding author e-mail: \tt\small jingyuso@umich.edu}
 }

\date{February 2024}

\newcommand{\tran}{\mathbf{p}}
\newcommand{\R}{\mathbf{R}}
\newcommand{\vel}{\mathbf{v}}
\newcommand{\bias}{\mathbf{b}}
\newcommand{\ba} {\bias^a} %
\newcommand{\bg} {\bias^g} %
\newcommand{\bv} {\bias^v} %
\newcommand{\SO}{\mathrm{SO}}
\newcommand{\SOthree}{\SO(3)}
\newcommand{\Real}{\mathbb{R}}
\newcommand{\Realthree}{\Real^{3}}

\renewcommand{\baselinestretch}{0.98}

\begin{document}

\maketitle

\begin{abstract}
Significant work has been done on advancing localization and mapping in underwater environments.
Still, state-of-the-art methods are challenged by low-texture environments, which is common for underwater settings. This makes it difficult to use existing methods in diverse, real-world scenes.
In this paper, we present TURTLMap, a novel solution that focuses on textureless underwater environments through a real-time localization and mapping method.
We show that this method is low-cost, and capable of tracking the robot accurately, while constructing a dense map of a low-textured environment in real-time.
We evaluate the proposed method using real-world data collected in an indoor water tank with a motion capture system and ground truth reference map. Qualitative and quantitative results validate the proposed system achieves accurate and robust localization and precise dense mapping, even when subject to wave conditions. The project page for TURTLMap is \url{https://umfieldrobotics.github.io/TURTLMap}.

\end{abstract}

\section{Introduction}
\label{Sec:introduction}

Robotic systems play an important role in improving our understanding of underwater environments~\cite{6174326, kinsey2006survey}. Among them, unmanned underwater vehicles (UUVs) stand out for their role in environmental monitoring and ocean exploration~\cite{kinsey2006survey, qin2023novel}. To safely conduct these tasks, UUVs are required to have accurate and robust localization (i.e., state estimation) capabilities~\cite{qin2023novel,rahman2019svin2, rahman2022svin2_rss, zhao2023tightly, song2023uncertainty}. However, achieving such accurate localization for UUVs is non-trivial due to the lack of GPS and Wi-Fi while the vehicle is underwater~\cite{rahman2019svin2, rahman2022svin2_rss}. %

The past few decades have seen great advances in localization systems, enabling robots on land to achieve real-time state estimation in complex environments~\cite{lin2023proprioceptive, yu2023fully}. 
These methods can be categorized into odometry or simultaneous localization and mapping (SLAM) techniques~\cite{lin2023proprioceptive,khattak2019keyframe, zhang2021new,ghaffari2019continuous ,murORB2, ORBSLAM3_TRO, labbe2019rtab}. With these methods, vision is usually combined with inertial measurements to improve accuracy and robustness~\cite{forster2016manifold}. 
However, underwater environments pose unique challenges to achieving similar performance for marine robotic systems compared to terrestrial systems equipped with visual or visual-inertial sensor configurations~\cite{iscar2017multi, rahman2019svin2, rahman2022svin2_rss, mallios2017underwater_caves, joshi2019experimental_comparison}. 
For vision-based localization algorithms, common challenges include visibility change, light and color attenuation, absence of light, and feature sparsity~\cite{fabbri2018enhancing, oliver2010image}. 
Despite these challenges, recent work has been successful in integrating vision into a robust localization system~\cite{kim2013real, zhao2023tightly}.
SVin2~\cite{rahman2019svin2,rahman2022svin2_rss} is one of the state-of-the-art (SOTA) underwater SLAM methods fusing camera, sonar, IMU and a barometer for improved accuracy and robustness. However, it still struggles in low-texture regions, which are often encountered in underwater environments. To tackle these challenges, additional modalities are usually introduced.
The Doppler velocity log (DVL) is a popular acoustic sensor supporting underwater relative navigation. 
Existing methods have seen success of fusing DVL, inertial, and pressure sensors for robust odometry with satisfying performance~\cite{song2023uncertainty, potokar2021invariant}. 
Xu et al.~\cite{xu2021underwater} fuse camera and DVL in a SLAM framework with improvement in low-texture environments, but they do not address dense mapping. Researchers in~\cite{wang2023real} propose a real-time dense mapping solution built on~\cite{rahman2019svin2, rahman2022svin2_rss}. However, the method's reliance on vision-based state estimation poses open challenges in general underwater environments with low-texture scenes.

\begin{figure}[t!]
    \centering
    \includegraphics[width=1.0\linewidth]{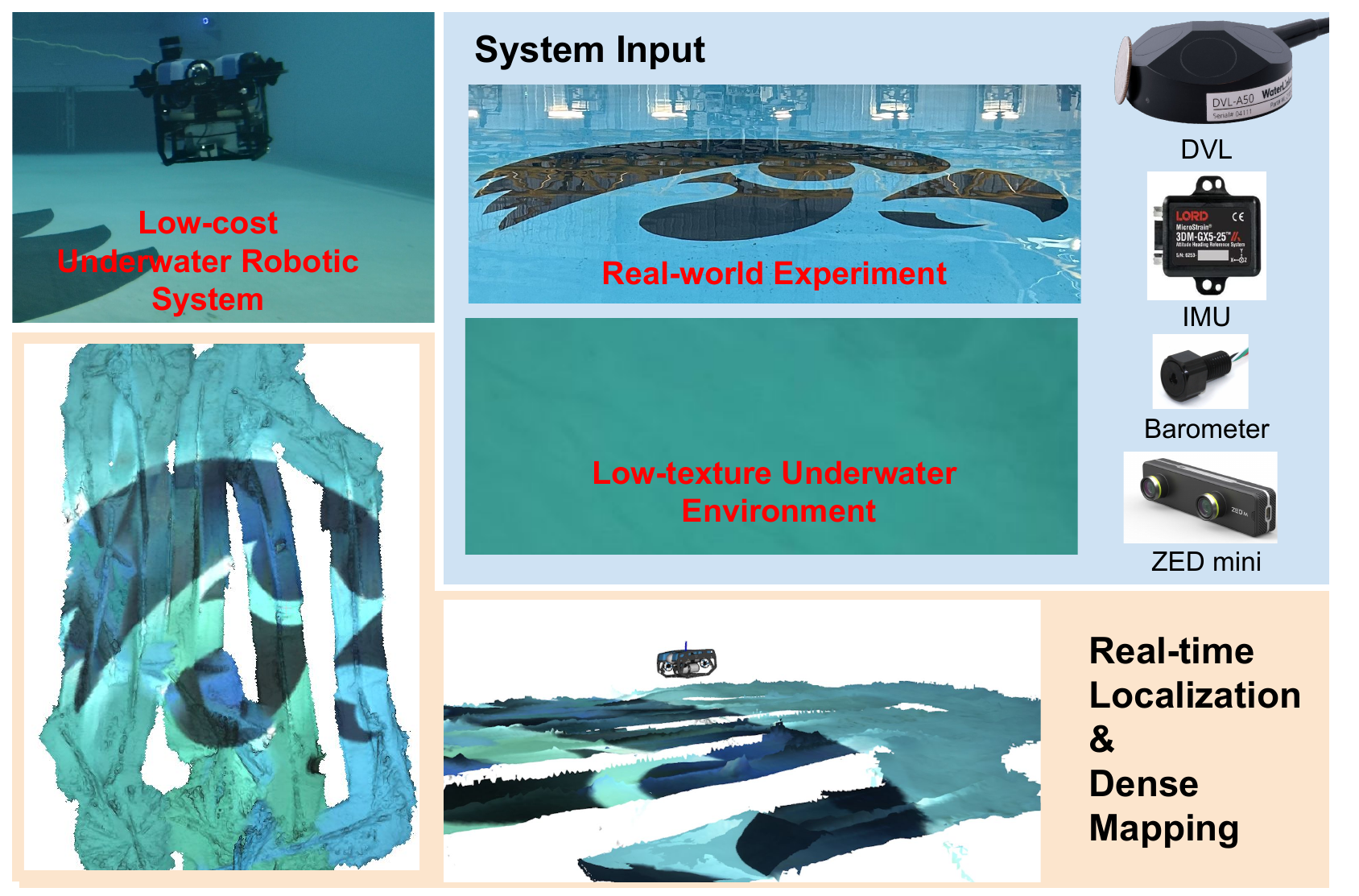}
    \caption{Overview of TURTLMap. We propose a real-time localization and dense mapping solution for low-texture underwater environments with a low-cost underwater robot.}
    \label{fig:pitch}
    \vspace{-6mm}
\end{figure}

In this work, we propose and demonstrate \textbf{TURTLMap}, a novel framework focusing on \textbf{T}extureless \textbf{U}nderwater environments through a \textbf{R}eal-\textbf{T}ime \textbf{L}ocalization and \textbf{Map}ping method (Fig.~\ref{fig:pitch}).
In these environments, visual sensors are usually unreliable for navigation. %
Our main contributions are: (i) We propose a pose graph-based DVL-inertial-barometer localization package that achieves improved localization accuracy in low-texture regions. (ii) With the proposed localization package and a real-time volumetric mapping package, we demonstrate real-time and accurate dense mapping on an onboard embedded computer on a low-cost UUV with a downward-facing stereo camera in a low-texture underwater environment. (iii) We provide extensive quantitative and qualitative evaluation on localization and mapping performance with ground truth localization provided by an underwater motion capture system and reference mapping provided by a 3D CAD model of the experimental site with and without wave conditions. Our contribution includes detailed description of integrated hardware and we provide open-source software supporting future implementation and extension for the research community, which will be made available here: \url{https://umfieldrobotics.github.io/TURTLMap}. %

\section{Related Works}

\subsection{Localization for Marine Robots}
Localization (i.e., state estimation) is one of the key tasks for underwater robots~\cite{rahman2019svin2, rahman2022svin2_rss, zhao2023tightly}. 
SLAM-based methods have been widely adopted to the underwater domain and they typically fuse vision, inertial sensors and/or acoustic sensors~\cite{kim2013real, xu2021underwater, rahman2019svin2, rahman2022svin2_rss}. 
The current SOTA underwater SLAM system, SVin2, presents a multi-sensor fusion approach that leverages digital pipe profiling sonar, visual and inertial sensors, and a barometer~\cite{rahman2019svin2, rahman2022svin2_rss}. 
SVin2 relies on feature detection from camera images for its visual front-end~\cite{khattak2019keyframe}. 
Though it fuses other modalities such as sonar and IMU, the localization module is still vision-oriented, making it hard to scale to featureless (low-texture) underwater environments~\cite{wang2023real, bharatswitch}. 

To reduce reliance on visual features for localization, it is common to integrate a DVL to provide seafloor-relative velocity. 
Researchers in~\cite{xu2021underwater} fuse vision with DVL odometry to achieve robust localization. Our method instead relies on DVL, IMU, and barometer sensors for localization for enhanced robustness in low-texture environments with wave conditions. We additionally integrate a vision-based mapping module capable of real-time dense reconstruction.
The sensor configuration in~\cite{zhao2023tightly} is similar to ours but in a reversed upward-looking orientation for under ice exploration. This method proposes fused odometry with camera, DVL, IMU and a pressure sensor for a robot navigating in a low-texture under ice environment. 
It differs from our method as it assumes planar features and slowly-varying altitude to aid visual tracking, which could be valid in its target scenario (i.e., under ice exploration) but does not apply to general underwater low-texture environments. 
Inspired by the success of recent DVL-based odometry methods~\cite{song2023uncertainty,potokar2021invariant,lin2023proprioceptive}, we design a pose graph-based state estimation framework fusing DVL, IMU and pressure sensors via GTSAM~\cite{gtsam}. 
Since our method does not require visual input for localization, there is no need for specific assumptions and aid for visual features, making it robust in low-texture environments with reliable localization accuracy. 
Another similar work~\cite{thoms2023tightly} formulates a DVL factor and fuses it with an existing LiDAR-camera SLAM algorithm for a surface vehicle doing maritime infrastructure inspection. Our method takes inspiration from its DVL factor design and further develops a real-time framework supporting additional sensors (i.e., barometer) for UUVs in an underwater setting.

\subsection{Mapping in Underwater Scenes}
There has been prominent work on underwater mapping, with works ranging from acoustic-based~\cite{song2023uncertainty}, large-scale mapping~\cite{uw_exploration}, to close-range visual mapping \cite{rahman2019svin2,rahman2022svin2_rss,johnson-roberson_high-resolution_2017, sethuraman2023waternerf, zhang2023beyond}. In this section, we review works relating to vision-based mapping, with an emphasis on real-time or dense mapping methods used in underwater robotics applications. %
Methods that enable real-time mapping in underwater environments typically utilize SLAM-based formulations \cite{okvis}, where the visual component plays a crucial role in both localization and mapping. In SVin2 \cite{rahman2019svin2, rahman2022svin2_rss}, a camera-based system is fused with a sonar in order to produce a feature-based map. The produced map is sparse, and relies on a feature-rich environment for denser and more accurate mapping. Wang et al.~\cite{wang2023real} leverage the localization output from SVin2 to develop a state-of-the-art method for real-time dense 3D mapping in underwater environments. However, experiments on real test sites demonstrate open challenges, showing that both localization and mapping performance suffers in environments that lack rich features, such as when the vehicle moves over a sandy region. Our proposed method overcomes this challenge by leveraging multi-modal sensors for robust localization in low-texture regions, which enables dense mapping with a real-time visual mapping framework. %
In particular, we leverage Voxblox~\cite{oleynikova2017voxblox} to achieve real-time dense 3D mapping following its successful application in other robotics fields~\cite{dang_graphbased_2020}.

\begin{figure*}[t!]
    \centering
    \includegraphics[width=1.0\linewidth]{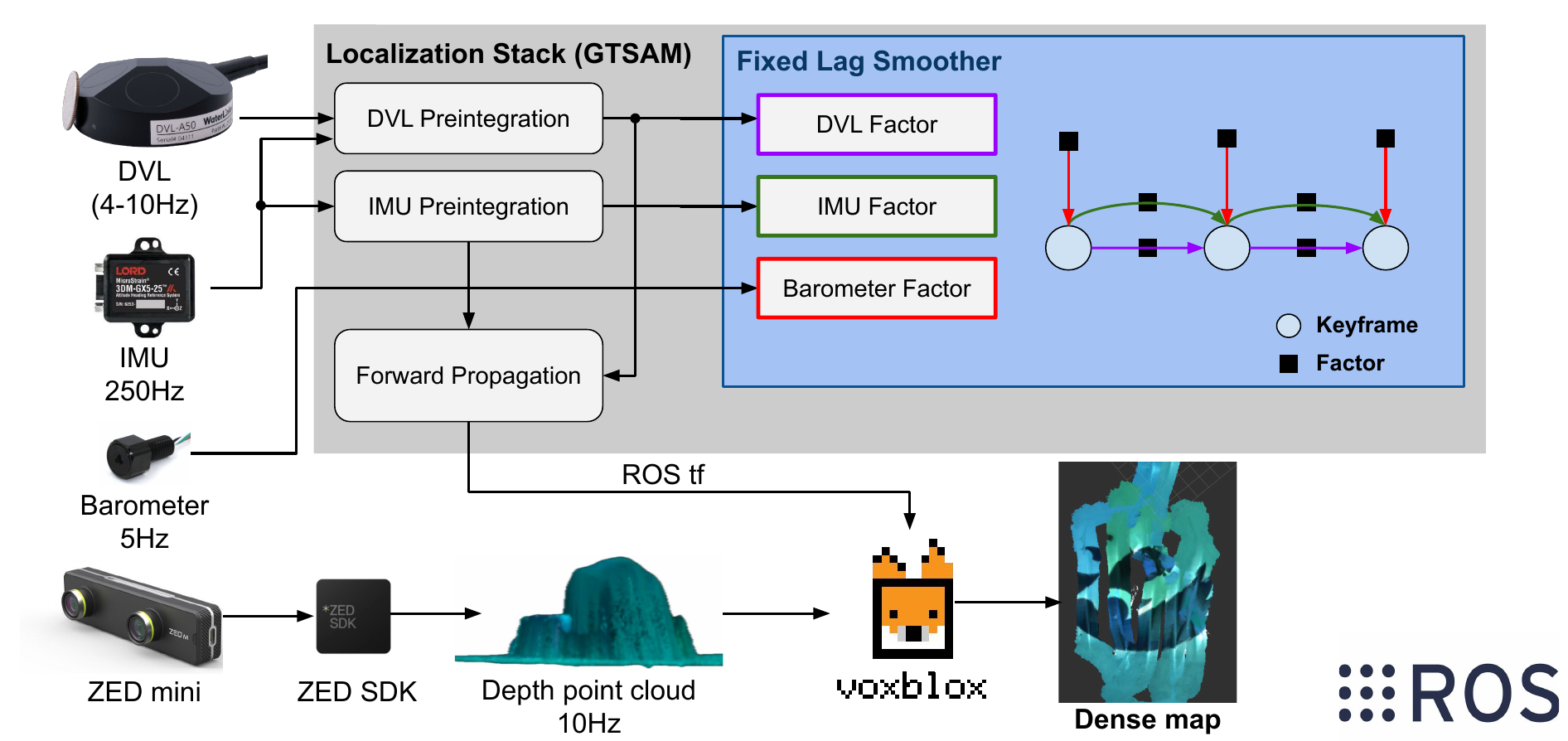}
    \caption{An overview of TURTLMap for real-time localization and dense mapping. We develop the system using ROS~\cite{ros}. The localization stack is formulated as a pose graph optimization framework with DVL, IMU and barometer measurements. The estimated pose is published as a transform message using ROS tf tool. Voxblox~\cite{oleynikova2017voxblox} subscribes to the pose and the depth point cloud from the ZED mini camera to construct a dense map in real-time.}
    \label{fig:overview_localization_mapping}
    \vspace{-4mm}
\end{figure*}

\section{Technical Approach}
Figure~\ref{fig:overview_localization_mapping} shows an overview of the proposed system for localization and mapping in low-texture underwater environments. TURTLMap takes the measurements from all navigation sensors (i.e., DVL, IMU and barometer) as input to the localization stack. The mapping module subscribes to the estimated pose and camera depth to conduct real-time dense mapping. Each component of the system will be discussed in more detail below.

\subsection{Sensor Integration}
Our underwater robotic system is equipped with a DVL, an IMU and a barometer for navigation, and a stereo camera for dense mapping. The DVL provides direct measurement of the linear velocity of the vehicle, while the IMU measures the angular velocity and acceleration. The barometer measures the depth of the vehicle to the water surface based on the water pressure. For mapping, the stereo camera provides dense depth as a point cloud. The measurements from the aforementioned sensors are handled in ROS~\cite{ros}. Detailed hardware setup will be introduced in Section~\ref{Sec:vehicle_setup}.

\subsection{Pose Graph Localization}
We formulate the state estimation problem as a factor graph with our integrated DVL, barometer, and IMU as navigation sensors. Specifically, we implement a keyframe-based maximum-a-posteriori DVL-inertial-barometer estimator in GTSAM~\cite{gtsam}. %
We follow the state-of-the-art Kimera-VIO~\cite{Rosinol20icra-Kimera} for framework design and add customized factors for the DVL and barometer, discussed in more detail below.

The state of the system at time $i$ can be described as:
\begin{equation}
    \mathbf{x}_i = [\R_i, \tran_i, \mathbf{v}_i, \bg_i, \ba_i, \bv_i].
\end{equation}
We assume that the IMU frame coincides with the body frame, and use the prior extrinsic calibration to handle measurements taken in different sensor frames. We define $\R_i \in \SOthree$ as the vehicle rotation, $\tran_i \in \Realthree$ is the vehicle position, $\bg_i, \ba_i \in \Realthree$ are the biases of IMU gyroscope and accelerometer measurements, and $\bv_i \in \Realthree$ is the linear velocity bias for the DVL. The rotation $\R_i$ and position $\tran_i$ are expressed in the world coordinate frame (NED convention), while other state variables are expressed in the body frame.
In our approach, we estimate the state of all keyframes:
\begin{equation}
    \mathcal{X}_k = {\{\mathbf{x}_i\}}_{i \in \mathcal{K}_k},
\end{equation}
where we denote $\mathcal{K}_k$ as the set of all keyframes up to time $k$. This design maintains a sparse pose graph that ensures fast runtime. We follow Kimera-VIO~\cite{Rosinol20icra-Kimera} and VILENS~\cite{wisth2022vilens} to use a fixed-lag smoother with a pre-defined time horizon. The fixed-lag smoother can help to bound the estimation time by marginalizing states that fall out of the smoothing horizon, which is important to ensure consistent real-time performance over long experimental duration.

We set up a ROS node to subscribe to different sensor drivers for their measurements. We assign an individual thread for IMU data since it is publishing at a high frequency (i.e., 250Hz). The IMU measurements are preintegrated using the implementation of~\cite{forster2016manifold} in GTSAM~\cite{gtsam}. We take inspiration from~\cite{wisth2022vilens} to design a preintegration function for DVL measurements. We also design a barometer factor to supervise the vehicle depth.

For every keyframe, we add a preintegrated IMU factor, a preintegrated DVL factor and a barometer factor. It is important to handle the sensor synchronization appropriately. Since IMU measurements have a high frame rate, every DVL and barometer measurement has a matched IMU measurement within tolerable time difference. In our sensor setup, the frame rate of the barometer is around 5 Hz, while the DVL frequency can vary between 4 and 15 Hz depending on the altitude. We insert a new keyframe based on the time elapsed from the last keyframe using the barometer message timestamp. This strategy ensures an accurate measurement for the vehicle depth for every keyframe, which fully leverages the advantage of the barometer.

Furthermore, as the DVL message frequency can be as low as 4 Hz, the time gap between the latest DVL measurement and the current keyframe is not negligible. By assuming the IMU propagation can estimate an accurate velocity change over a short time horizon (i.e., max 0.25 second), we propose to use the IMU preintegration to obtain the latest velocity for the current keyframe and add that as a pseudo measurement to the DVL preintegration.

\subsubsection{IMU Factor}
We follow the existing works~\cite{Rosinol20icra-Kimera, wisth2022vilens} to use the preintegrated IMU factor to constrain the pose and velocity between two consecutive keyframes. This idea is first proposed in~\cite{forster2016manifold} to remove the need for recomputation of the integration between two keyframes. We use the same preintegration design implemented in GTSAM~\cite{gtsam} to handle the high frequency signals. The parameters (e.g., noise density, noise random walk) used in IMU preintegration are set according to the data sheet of the IMU. As mentioned above, we also set up an additional IMU preintegration for estimating the velocity change since the latest DVL measurement. This additional preintegration is reset every time a new DVL measurement is received. We refer to~\cite{forster2016manifold} for more details about preintegrated IMU factors.

\subsubsection{Barometer Factor}
The barometer provides accurate depth measurements based on the water pressure. Since the robot state is defined in the NED frame, the depth of the robot at time $i$ is defined as
\begin{equation}
    d_i = {p^z}_i - {p^z}_0,
\end{equation}
where $d_i$ is the depth compensated by the initial depth, and $p^z$ is the $z$ axis value of $\tran$.
For the depth measurement ${z}_i$, we define the residual as
\begin{equation}
    r_{d_i, {z}_i} = {d}_i - {z}_i.
\end{equation}
The barometer is integrated as a unary factor on $\tran_i$, and its Jacobian used in our factor implementation is
\begin{equation}
    \frac{\partial{r_{{d}_i, {z}_i}}}{\partial{\delta{\tran_i}}} = \begin{bmatrix}
0 & 0 & 1
\end{bmatrix} \times \R_i,
\end{equation}
which is a $1\times3$ vector derived by following the lifting strategy introduced in~\cite{forster2016manifold}.

\subsubsection{DVL Factor}
The DVL measures the linear velocity of the robot. We follow common practices~\cite{thoms2023tightly, wisth2022vilens, song2023uncertainty} to use the constant velocity model for velocity integration. Unlike~\cite{wisth2022vilens, thoms2023tightly}, we decouple the IMU preintegration with the DVL preintegration and use the estimated altitude change from the IMU's onboard Extended Kalman Filter (EKF). According to our analysis and comparison with ground truth, the estimated altitude change from the onboard EKF of our IMU is accurate in the short time gap between consecutive DVL measurements. In addition, decoupling the IMU preintegration reduces the system computation load and can leverage the fine-tuned onboard attitude reference system. We also directly leverage the reported DVL velocity measurement uncertainty from the manufacturer driver to inform the DVL factor of the measurement quality. As we observe that the uncertainty of DVL measurements can increase considerably due to multiple reasons (e.g., strong environmental disturbance such as waves, fast change of altitude in scenarios such as driving from beach to sea, out of operating range), this design has advantages of tuning the weight for DVL factors in the pose graph adaptively. 

We follow~\cite{song2023uncertainty,wisth2022vilens} to add a slowly varying bias term $\bv$ to the velocity:
\begin{equation}
    \Tilde{\vel} = \vel+\bv+\bm{\eta}^{v},
\end{equation}
where $\eta^{v}$ is the linear velocity Gaussian white noise term. We formulate the DVL factor as a translation residual between two consecutive keyframes. The translation measurement is defined as:
\begin{equation}
    \Delta \Tilde{\tran}_{ij} = \sum _{k=i}^{j-1} \Delta \tilde{\R}_{ik}(\tilde{\vel}_k - \bv_i) \Delta t,
\end{equation}
where $k$ is an intermediary measurement between keyframes $i$ and $j$. As mentioned above, we decouple the IMU preintegration by directly using the $\Delta \tilde{\R}_{ik}$ measurement from the IMU's onboard EKF. Our proposed DVL factor residual is formulated as
\begin{equation}
    \mathbf{r}_{\mathcal{D}_{ij}} = \R_i^\intercal(\tran_j - \tran_i) - \Delta \Tilde{\tran}_{ij}(\bv_i).
\end{equation}

The Jacobians for the involved state variables $\tran_i$ and $\tran_j$ can be derived following~\cite{forster2016manifold}. We follow~\cite{thoms2023tightly, wisth2022vilens} to implement the Jacobian for $\bv_i$. We compute the covariance of $\mathbf{r}_{\mathcal{D}_{ij}}$ iteratively,
\begin{equation}
    \mathbf{\Sigma}^\mathcal{D}_{i,k+1} = \mathbf{A} \mathbf{\Sigma}^\mathcal{D}_{i,k} \mathbf{A}^\top + \mathbf{B} \mathbf{\Sigma}^{\mathcal{D}}_\eta \mathbf{B}^\top,
\end{equation}
where the first term $\mathbf{\Sigma}^\mathcal{D}_{i,i}$ starts from $\mathbf{0}$, and $\mathbf{\Sigma}^{\mathcal{D}}_\eta$ is set as the measurement uncertainty from the DVL driver report. We derive $\mathbf{A} = \mathbf{I_{3\times3}}$ and $\mathbf{B}=\Delta \tilde{\R}_{ik} \Delta t$. We refer to~\cite{thoms2023tightly, wisth2022vilens} for a more detailed derivation process.

\subsection{Real-time Volumetric Mapping}
We leverage Voxblox~\cite{oleynikova2017voxblox}, a real-time volumetric mapping library that can run on embedded devices as the mapping algorithm of this work. Benefiting from its tight ROS integration, the integration of Voxblox with the proposed localization package is straight-forward. As shown in Fig.~\ref{fig:overview_localization_mapping}, Voxblox directly subscribes to the depth measurements of the ZED mini camera. It leverages the ROS tf tool to query the estimated pose of the camera output by our robust localization system to reconstruct the scene using the mesh representation.

\begin{figure}[h]
   \centering
    \includegraphics[width=1.0\linewidth]{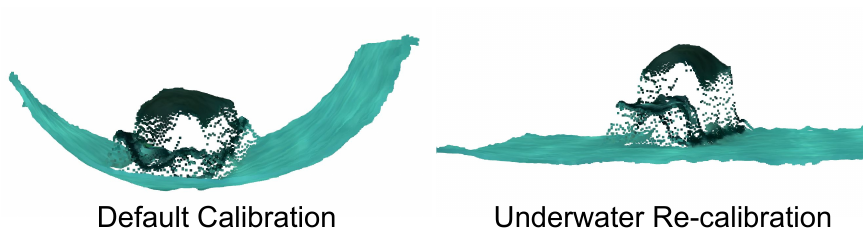}
   \caption{A comparison of the stereo depth point cloud of a rock platform placed on the tank bottom with the default calibration (left) and the underwater calibration (right). The bottom surface of the tank is flat.}
    \label{fig:calibration_comparison}
    \vspace{-4mm}
\end{figure}

We use the neural mode of the depth estimation in the ZED SDK, which is the most accurate and stable depth sensing mode according to our analysis. In our experiment, we set the camera resolution to VGA and the publishing rate of the depth point cloud to 10 Hz. These settings can be adjusted according to the computation power of the embedded device. One of the important factors affecting the mapping performance is the camera calibration. We re-calibrate the stereo camera for underwater usage. We use the multiple camera calibration tool in Kalibr toolbox~\cite{kalibr_camera} and a customized Aprilgrid target to obtain the underwater calibration parameters. We demonstrate a comparison of the depth point cloud between default calibration and the underwater calibration in Fig.~\ref{fig:calibration_comparison}. Note that the tank bottom should appear to be flat, so this validates the calibration using prior knowledge of the reference structure.

\section{Experiment \& Results}
\subsection{Vehicle Setup}
\label{Sec:vehicle_setup}
We set up the proposed TURTLMap package based on a low-cost robotic system with our proposed sensor configuration as discussed below.

\subsubsection{Robot Platform}
\begin{figure}
    \centering
    \includegraphics[width=0.96\linewidth]{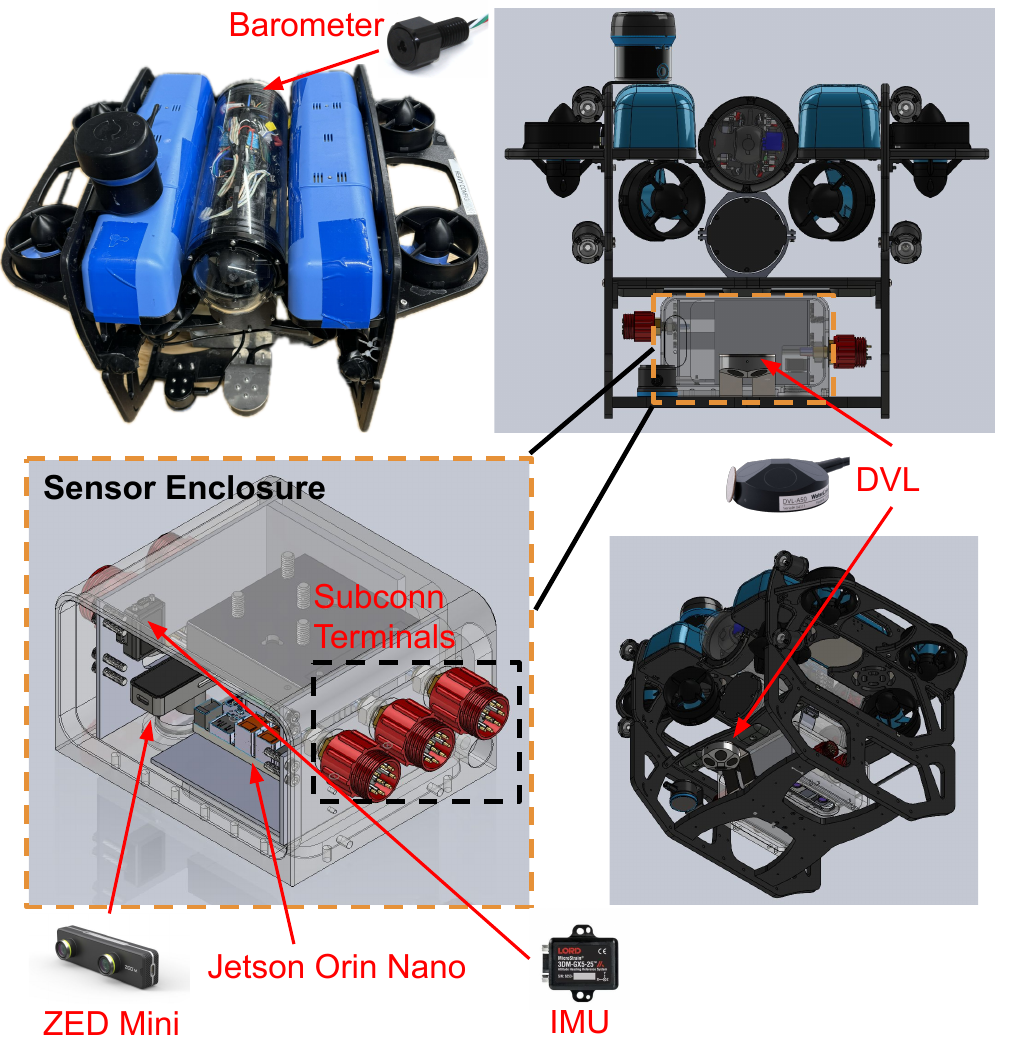}
    \caption{An overview of the robot platform and the customized sensor enclosure. The Blue Robotics Bar30 barometer is located in the electronics enclosure of the BlueROV2. The Waterlinked DVL A50 is placed on the payload skid with its transducers facing downward. The ZED mini stereo camera is placed inside the customized sensor enclosure with an NVIDIA Jetson Orin Nano embedded computer. We also place a LORD MicroStrain 3DM-GX5-25 IMU inside the sensor enclosure. Additional sensors (e.g., Blue Robotics Ping360 scanning sonar, Blue Robotics Ping1D sonar) are mounted on the robot, but they are not used in this work.}
    \label{fig:hardware_overview}
    \vspace{-7mm}
\end{figure}

We build the proposed solution based on the heavy configuration of the Blue Robotics BlueROV2 UUV, which is an affordable, high-performance ROV with high flexibility of customization. We use the official payload skid to accommodate the added hardware items. To accommodate the components that are not waterproof, we include a customized Sexton aluminum enclosure and mount it onto the payload skid of the BlueROV2. The proposed localization and mapping package is running on the NVIDIA Jetson Orin Nano embedded computer. Figure~\ref{fig:hardware_overview} shows an overview of the robot platform with the customized sensor enclosure.

\subsubsection{Hardware Configuration}
When choosing among various navigation and perception sensors, as well as the embedded computer, the most important consideration factors are the cost, size, and difficulty of integration. We choose to use the combination of DVL, IMU and barometer to build the localization stack. We choose the Waterlinked DVL A50 model due to it being a low-cost option among existing DVLs on the market, along with its compact size. Another advantage of this DVL is the official integration guide and support for the BlueROV2. Though the DVL is equipped with an internal IMU, its data is inaccessible. We only use the velocity measurements from the DVL to maintain compatibility of the proposed method to various DVL models. Though the BlueROV2 comes with an IMU, its frame rate is limited by MAVLink (10Hz by default), which is a major challenge for robust state estimation. We add an additional LORD MicroStrain 3DM-GX5-25 IMU to estimate the rotation of the vehicle. This IMU is reasonably priced and has an onboard EKF for accurate attitude measurement. The robot also comes with a Blue Robotics Bar30 barometer that has absolute measurement for robot depth. We access its measurement through communication with the BlueROV2 Raspberry Pi via MAVLink for the proposed state estimation package.

Though pose tracking via visual sensors is not reliable in underwater low-texture environments, we value its importance in dense reconstruction of these environments (aided by robust localization). We choose the ZED mini stereo camera due to its reasonable price and small form factor that fit well to our budget and limited payload space. We place the ZED mini camera downward-facing in the custom waterproof enclosure. The ZED mini camera is accompanied by the ZED SDK, providing useful functions for robotic application (e.g., image capturing, depth sensing). In this work, we use the depth measurements from the ZED SDK for mapping. It is worth mentioning that the IMU inside the ZED mini camera is not used in this project due to its data quality and low frame rate. The ZED SDK requires an embedded computer with a GPU. We choose to use an NVIDIA Jetson Orin Nano due to the low cost. %
It also has a compact design that can fit in the sensor enclosure easily and we have tested that it has stable thermal performance when placed in an enclosure.

We maintain proper calibration to ensure accurate localization and mapping. We use the IMU-camera calibration function in Kalibr toolbox~\cite{kalibr_1, kalibr_2} to obtain the transformation between the IMU and the ZED mini stereo camera. We leverage the CAD model of the robot platform to obtain the extrinsics for other sensors.

\begin{figure}[t]
    \centering
    \includegraphics[width=1.0\linewidth]{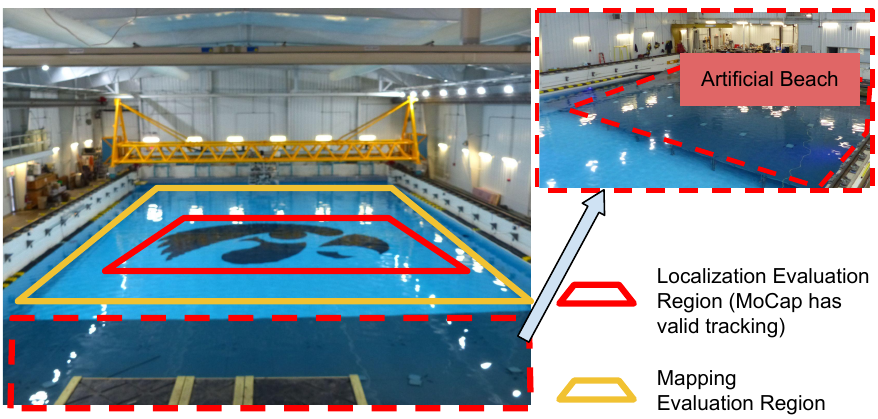}
    \caption{The towing tank at the University of Iowa. We show different regions for localization evaluation and mapping evaluation. The region where the MoCap system has valid tracking is relatively small. We exclude the artificial beach for map evaluation due to the lack of its reference structure.}
    \label{fig:tank_structure}
    \vspace{-3.2mm}
\end{figure}

\begin{table}[tp!]
    \centering
    \caption{Details of the logs used in the evaluation of the proposed method.}
    \begin{tabular}{c|ccc}
        \hline
        Log & Wave & Description & Duration (s) \\ \hline 1   & \xmark    & \begin{tabular}[c]{@{}c@{}}Robot surveys the tank \\ while being maintained \\ about 1.75 m to the bottom\end{tabular} & 352 \\ \hline 2   & \cmark    & \begin{tabular}[c]{@{}c@{}}Wave Height: 0.1 m \\ Wave Frequency: 1 Hz\\ Robot drives two squares and \\maintains at the water surface\end{tabular}& 280
    \end{tabular} 
    \label{tab:driving_pattern}
    \vspace{-7.5mm}
\end{table}

\begin{figure}[b]
    \centering
    \includegraphics[width=0.98\linewidth]{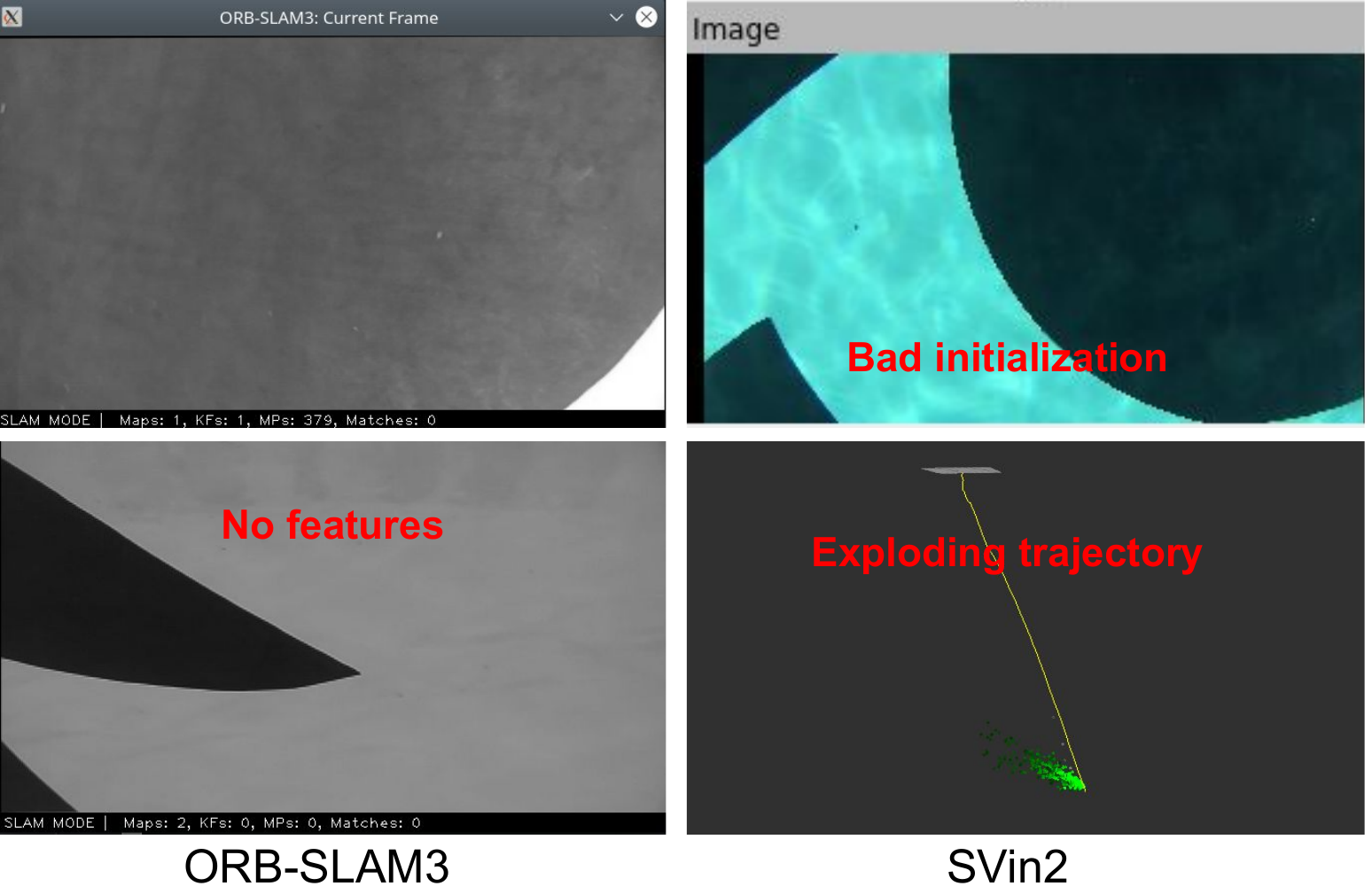}
    \caption{Low-texture underwater environments are challenging to state-of-the-art visual localization methods (i.e., ORB-SLAM3~\cite{ORBSLAM3_TRO}, Svin2~\cite{rahman2019svin2, rahman2022svin2_rss}).}
    \label{fig:vslam_failure}
    \vspace{-1mm}
\end{figure}

\begin{table}[t!]
    \centering
    \caption{Comparison of the localization accuracy of the proposed method~(TURTLMap) and baselines. Abbreviations: DVL (D), IMU (I), Barometer (B). $\text{ATE}_\text{pos}$ is in meters and $\text{ATE}_\text{rot}$ is in degrees.}
    \begin{tabular}{c|ccc|cc|cc}
\hline
\multirow{2}{*}{Method} & \multicolumn{3}{c|}{Modality} & \multicolumn{2}{c|}{Log 1}  & \multicolumn{2}{c}{Log 2}   \\ \cline{2-8} 
                        & D        & I        & B       & $\text{ATE}_\text{pos}$ & $\text{ATE}_\text{rot}$ & $\text{ATE}_\text{pos}$ & $\text{ATE}_\text{rot}$ \\ \hline
DVL Odom.                & \cmark   & \cmark   & \xmark  & 1.84        & 7.52          & 0.99        & 6.52          \\
UKF~\cite{song2023uncertainty}                     & \cmark   & \cmark   & \cmark  & 0.48        & 11.86         & 0.36        & 15.8          \\
TURTLMap                    & \cmark   & \cmark   & \cmark  & \textbf{0.18}        & \textbf{3.72}          & \textbf{0.26}        & \textbf{4.22}         
\end{tabular}
    \label{tab:quantitative_localization}
    \vspace{-2mm}
\end{table}

\subsection{Experiment Setup}
To demonstrate the capability of the proposed framework, experiments were conducted in the towing tank at the University of Iowa. We use the wave generator to create natural environment disturbance to mimic real-world challenges for localization. The towing tank is an ideal place to evaluate the proposed system due to the equipped underwater motion capture (MoCap) system that can be used as the ground truth reference for vehicle trajectory, and a reference structure of the tank for validating the mapping accuracy. We show an overview of the tank in Fig.~\ref{fig:tank_structure}, which also indicates regions we use for evaluating the localization and mapping modules.

\begin{figure}[tbp!]
    \centering
    \includegraphics[width=0.95\linewidth]{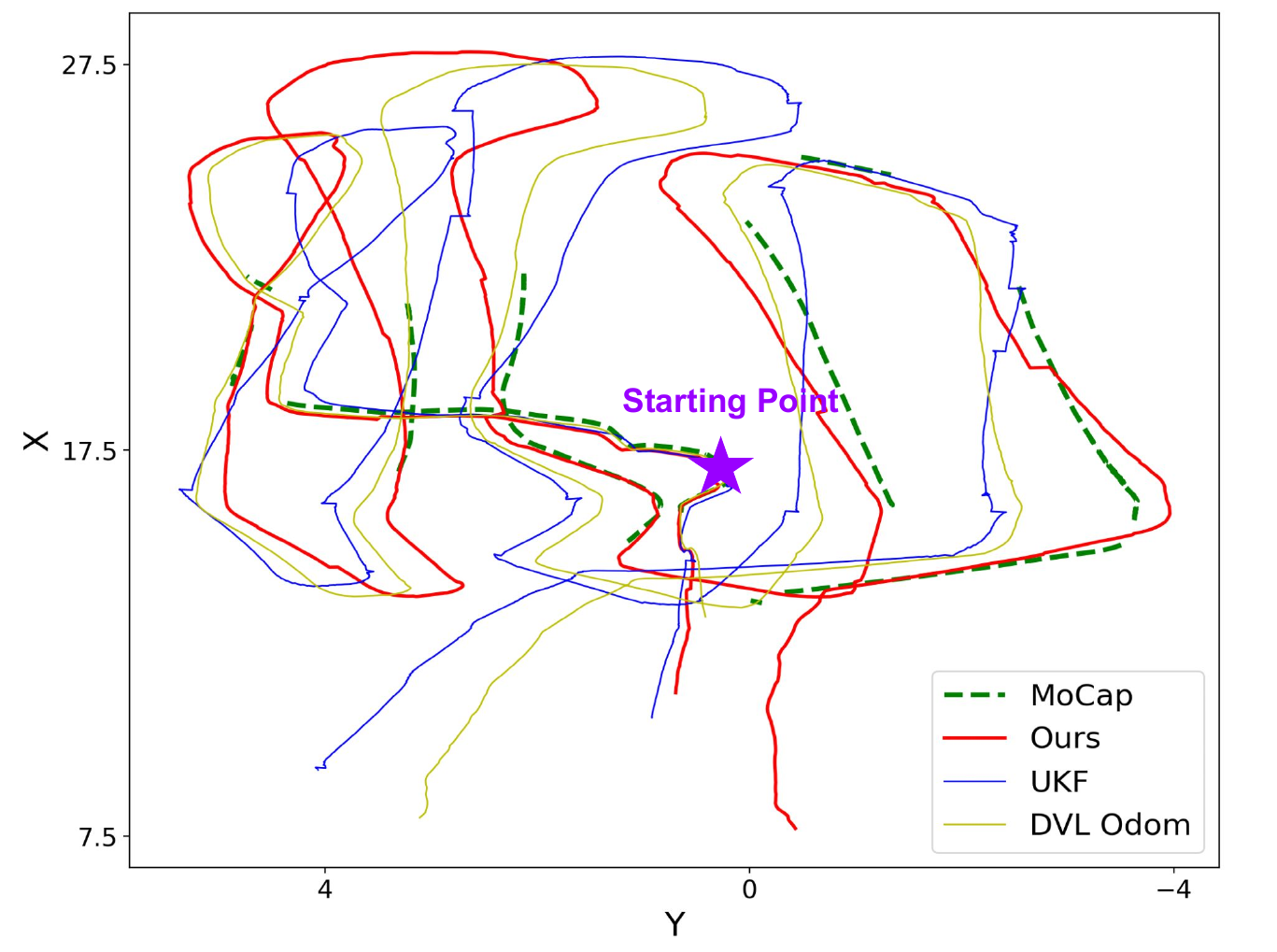}
    \caption{A qualitative comparison of estimated trajectories for log 1 in the x-y plane shown in meters. The starting point indicates where the robot enters the MoCap's valid tracking zone. Our method maintains the most accurate trajectory with the least amount of drift over time. Best viewed with color and zoomed-in.}
    \label{fig:localization_trajectory}
    \vspace{-7mm}
\end{figure}

\begin{figure*}[t]
    \centering
    \includegraphics[width=0.95\linewidth]{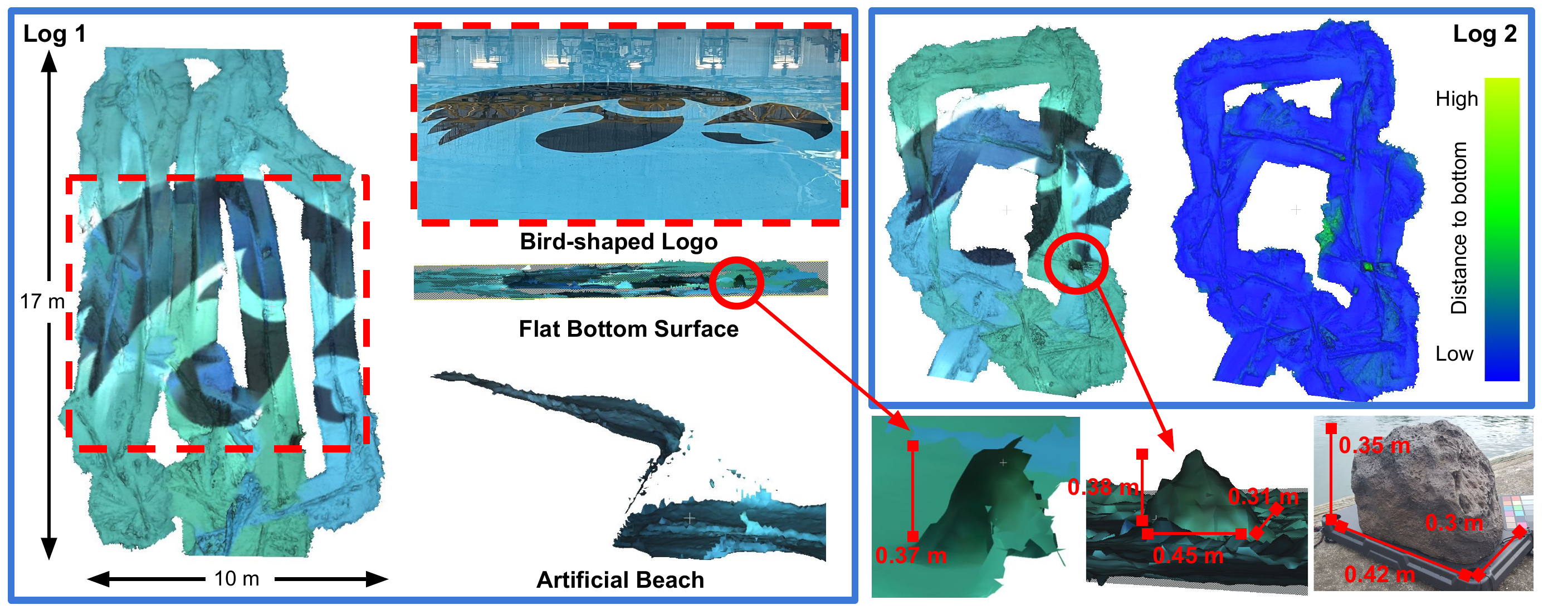}
    \caption{Qualitative results of the map generated from both logs. In log 1, we highlight the reconstructed bird-shaped logo and the corresponding image. The map matches the flat bottom surface well while including details of the artificial beach. In log 2, we plot a color map of distance to the bottom surface showing the capability of the proposed solution under wave disturbance. We also show a comparison of the rock dimensions captured from both logs.}
    \label{fig:qualitative_mapping}
    \vspace{-6mm}
\end{figure*}

\subsection{Evaluation Setting}
We use two logs collected at different wave and driving patterns as listed in Table~\ref{tab:driving_pattern}. To evaluate localization performance, we use the the open-source tool~\cite{Zhang18eval} to evaluate absolute trajectory error (ATE) between the output of our localization algorithm and the MoCap trajectory. As noted in Fig.~\ref{fig:tank_structure}, the MoCap system does not have full coverage of the tank, so we only conduct evaluation in the region where the MoCap system has valid tracking of the robot. As each log is initiated at the artificial beach region, which is outside of the MoCap tracking zone, we align the reference frame of our localization package to the MoCap reference frame according to the first valid MoCap pose. We run state-of-the-art (SOTA) vision-based SLAM methods, ORB-SLAM3~\cite{ORBSLAM3_TRO} and SVin2~\cite{rahman2019svin2, rahman2022svin2_rss}, as baselines to show the challenges that vision-based localization methods face in low-texture underwater environments. We further compare our proposed localization method against the onboard DVL dead reckoning results (DVL Odom.), and a baseline Unscented Kalman Filter (UKF) implemented in~\cite{song2023uncertainty}. %

We use the ground truth mesh reference from the CAD model to evaluate the generated dense map in Cloud Compare~\cite{cloudcompare}. Though the artificial beach is captured in our logs, due to the lack of the actual reference structure of the beach, we only evaluate the map with the bottom surface of the tank, as indicated in Fig.~\ref{fig:tank_structure}.
We also put an artificial rock platform with known 3D structure in the tank to validate the capability of the proposed solution in mapping small objects. In our experiments, we follow the default configuration of Voxblox~\cite{oleynikova2017voxblox} setting the voxel size to 0.05 m.

\subsection{Localization Results}
The underwater low-texture environments pose a significant challenge to the selected SOTA SLAM methods. ORB-SLAM3~\cite{ORBSLAM3_TRO} has a hard time passing the initialization stage with its stereo or stereo-inertial mode. Though we try to tune the related parameters and apply CLAHE image pre-processing following~\cite{rahman2019svin2, rahman2022svin2_rss}, due to the limited number of tracked features, the trajectory quickly diverges and the map is reset. A similar issue happens with SVin2~\cite{rahman2019svin2, rahman2022svin2_rss}, as we also notice the quickly exploding trajectory. We show examples of difficulties with these two baselines in Fig.~\ref{fig:vslam_failure}. This demonstrates that SOTA visual SLAM systems are not reliable for navigation in low-texture underwater environments.

We report quantitative comparison of our method and other localization baselines in Table~\ref{tab:quantitative_localization}. From this table, we can see that the proposed method is able to maintain the lowest ATE in both position and rotation. 
In addition, we plot the trajectories of the proposed state estimation algorithm and other baselines in Fig.~\ref{fig:localization_trajectory}, which shows the proposed method closely aligns to the ground truth trajectory with the least amount of drift. 

\begin{table}[t]
    \centering
    \caption{Ablation study of the proposed localization module in TURTLMap. $\text{ATE}_\text{pos}$ is in meters and $\text{ATE}_\text{rot}$ is in degrees.}
    \begin{tabular}{c|c|cc|cc}
\hline
\multirow{2}{*}{Pseodo Vel.} & \multirow{2}{*}{Sensor Unc.} & \multicolumn{2}{c|}{Log 1}                        & \multicolumn{2}{c}{Log 2}                          \\ \cline{3-6} 
                             &                              & $\text{ATE}_\text{pos}$ & $\text{ATE}_\text{rot}$ & $\text{ATE}_\text{pos})$ & $\text{ATE}_\text{rot}$ \\ \hline
\xmark                       & \cmark                       & 0.32                    & 5.44                    & 0.40                     & 4.72                    \\
\cmark                       & \xmark                       & 0.23                    & \textbf{3.63}                    & 0.28                     & 4.42                    \\
\cmark                       & \cmark                       & \textbf{0.18}                    & 3.72                    & \textbf{0.26}                     & \textbf{4.22}                   
\end{tabular}
    \label{tab:ablation}
    \vspace{-6mm}
\end{table}
In our localization system design, we propose to use IMU preintegration to generate an additional pseudo DVL velocity measurement to mitigate the time gap between the latest DVL measurement and current keyframe. We also design the DVL factor to be aware of the reported DVL velocity measurement uncertainty. We conduct an ablation study of these two design components. As shown in Table~\ref{tab:ablation}, using the IMU to predict a pseudo velocity measurement for the latest keyframe is effective in improving the localization accuracy. When compared with using a constant uncertainty model for the DVL velocity measurement, we see improvement from the proposed method as it fuses the sensor measurement quality reported from the manufacturer driver to determine the factor weight in the pose graph.

\subsection{Mapping Results}
We evaluate the dense maps generated by TURTLMap of the two logs in Table~\ref{tab:driving_pattern}. %
We compare it with the reference structure of the tank, which is a simple flat surface. Table~\ref{tab:quantitative_mapping} reports the error between the generated map and the ground truth reference with different localization methods as input. We do not include ORB-SLAM3~\cite{ORBSLAM3_TRO} and SVin2~\cite{rahman2019svin2, rahman2022svin2_rss} as they fail to generate a map. It is demonstrated that the proposed mapping system with our localization input is able to maintain an accurate map with low error compared to other localization baselines as our method produces more accurate state estimation. The comparison between the two logs also demonstrates that the environmental disturbance caused by waves introduces challenges to the proposed system, but the mapping error only increases to 0.04 m.

\begin{table}[t]
    \centering
    \caption{Quantitative results of mapping evaluation. }
    \begin{tabular}{c|cc}
\hline
\multirow{2}{*}{Localization Input} & \multicolumn{2}{c}{Mapping Error (m)} \\
                              & Log 1             & Log 2             \\ \hline
DVL Odom                      & 1.733             & 0.806             \\
UKF~\cite{song2023uncertainty}                           & 0.105             & 0.112             \\
TURTLMap (Ours)                          & \textbf{0.019}             & \textbf{0.040}            
\end{tabular}
    \label{tab:quantitative_mapping}
    \vspace{-7mm}
\end{table}

We also show a comparison of the generated map from different perspectives with images of the tank structure in Fig.~\ref{fig:qualitative_mapping}.
The qualitative comparison highlights the capability of the proposed real-time mapping solution. In log 1, the large-scale bird-shaped logo is reconstructed in detail in the generated map and matches the picture well, demonstrating the effectiveness of the proposed solution. The map generated from log 2 also validates the system's capabilities under wave disturbance, with the rock platform being mapped precisely. We show a color map with distance to the ground truth reference of the bottom surface to highlight the mapping accuracy. We additionally show a qualitative example of the artificial beach in the generated map, demonstrating the capability of the system in maintaining an accurate map in scenarios of significant altitude change. This capability is very important for safe navigation for UUVs.

\subsection{Runtime}
We measure the runtime of the localization module and mapping module on the Jetson Orin Nano embedded computer. The time cost of pose graph optimization at each keyframe varies between 0.01s and 0.04s. Due to the applied fixed lag smoother, the time cost is consistent over the complete run.
The time cost for each map update is consistently around 0.02s in the longest survey log. As these two modules are running on different threads, we can conclude that the proposed system is capable of running in real-time.

\section{Conclusion \& Future Work}
We have proposed TURTLMap, an effective real-time localization and dense mapping solution for a low-cost UUV. The robustness of TURTLMap is evaluated in challenging low-texture underwater scenes with and without waves. We develop a multi-modal localization stack using a factor graph. We conduct real-time volumetric mapping based on stereo depth information and estimated robot pose from the proposed localization module. We conduct extensive evaluation of the proposed system with real-world data including underwater MoCap and ground truth mapping reference, which shows the effectiveness of the proposed system.

In this work, only the stereo point cloud is used for mapping. Extending this work to fuse the sparse acoustic measurements for mapping is an interesting future direction. Another future direction is to evaluate the proposed system in real field environments with a mix of low-texture and high-texture scenes. In addition, we plan to extend the proposed system to add visual odometry factors and visual loop closure in environments with sufficient visual features, enhancing performance in activities such as shipwreck survey and coral reef monitoring~\cite{wang2023real,rahman2019svin2, rahman2022svin2_rss,sethuraman2024machine}. This will be a natural extension due to our factor graph formulation. %

\section{Acknowledgement}
We would like to thank Prof. Casey Harwood and Mike Swafford for their support with experiments at the University of Iowa. We thank Morgan Sun for integrating the embedded device. This work relates to Department of Navy award N00014-21-1-2149 issued by the Office of Naval Research.

\bibliographystyle{IEEEtran}
\bibliography{reference}

\end{document}